%% file: emnlp2021.tex
\pdfoutput=1

\documentclass[11pt]{article}
\usepackage{multirow}
\usepackage[table,xcdraw]{xcolor}
\usepackage{graphicx}
\usepackage{colortbl}
\usepackage{adjustbox}
\usepackage{emnlp2021}
\usepackage{subcaption}
\usepackage{times}
\usepackage{latexsym}
\definecolor{bananamania}{rgb}{0.98, 0.91, 0.71}
\usepackage[T1]{fontenc}
\usepackage[utf8]{inputenc}

\usepackage{microtype}
\usepackage{booktabs}
\usepackage{afterpage}
\usepackage{xspace}
\newcommand{\crapbank}{{\sc Fsmb}\xspace}
\newcommand\narabizi{{\em NArabizi}\xspace}
\newcommand{\modeltask}{{\sc Model+Task}\xspace}

\usepackage{morefloats}
\input{draft_final}

\newcommand\blfootnote[1]{%
  \begingroup
  \renewcommand\thefootnote{}\footnote{#1}%
  \addtocounter{footnote}{-1}%
  \endgroup
}

\title{Can Character-based Language Models Improve Downstream Task Performance \draftreplace{In}{in} Low-Resource \draftreplace{And}{and} Noisy Language Scenarios?}

\author{Arij Riabi$^{1,2}$
  \quad
  Benoît Sagot$^{1}$ 
  \quad 
  Djamé Seddah$^{1}$\\
  $^1$ Inria Paris\\
  $^2$ Sorbonne Universit\'e\\
  {\tt \{arij.riabi,benoit.sagot,djame.seddah\}@inria.fr} \\}

\begin{document}
\maketitle
\begin{abstract}

  Recent impressive improvements in NLP, largely based on the success
  of contextual neural language models, have been mostly demonstrated
  on at most \draftreplace{20}{a couple dozen} high-resource languages. 
Building language models and, more generally,
  NLP systems for non-standardized and low-resource languages remains
  a challenging task. In this work, we focus on North-African
  colloquial dialectal Arabic written using an extension of the Latin
  script, called \textit{NArabizi}, found mostly on social media and
  messaging communication. In this low-resource scenario with data
  displaying a high level of variability, we compare the downstream performance
  of a character-based language model on part-of-speech tagging and
  dependency parsing to that of monolingual and multilingual models.
  We show that a character-based model trained on only 99k sentences
  of \textit{NArabizi} and fined-tuned on a small treebank of this
  language leads to performance close to those obtained with the same
  architecture pre-trained on large multilingual and monolingual
  models. Confirming these results a on much larger data set of noisy
  French user-generated content, we argue that such character-based
  language models can be an asset for NLP in low-resource and high
   language variability settings.  
\end{abstract}

\section{Introduction}\blfootnote{First version submitted on August
  27th, 2021. Final on October 1st, 2021.}

Current state-of-the-art monolingual and multilingual language models
require large amounts of data to be trained, showing limited
performance on low-resource languages \cite{howard2018universal,
  devlin-etal-2019-bert}. They lead to state-of-the-art results on
most NLP tasks \cite{devlin2018bert,raffel2019exploring}. In order to
achieve high performance, these models rely on transfer learning
architectures: the language models need to be trained on large amounts
of data ({\em pre-training}) to be able to transfer the acquired
knowledge to a downstream task via {\em fine-tuning} on a relatively
small number of examples, resulting in a significant performance
improvement with respect to previous approaches. This dependency on
large data sets for pre-training is a severe issue for low-resource
languages, despite the emergence of large and successful multilingual
pre-trained language models \cite{muller-etal-2021-first}. This is
especially the case for languages with unusual morphological and
structural features, which struggle to take advantage from
similarities with high-resource, well represented languages such as
Romance and Germanic languages.

\iftrue In this work, we \draftremove{choose to} focus on one of such
highly challenging languages, namely North-African dialectal Arabic.
Its Latin transcription (\emph{Arabizi}) displays a high level of linguistic
variability\footnote{Language variability, or language
   variation, is  a term coming from socio-linguistics where, as
  stated by \newcite{Nordquist:2019:variability}, it {\em refers
   to regional, social or contextual differences in the ways that a
   particular language is used}. These variations in user-generated content can be characterized
through their prevalent idiosyncraisies when compared to canonical
texts \cite{seddah-etal-2012-french,DBLP:journals/corr/abs-2011-02063}.}, \draftreplace{with}{on top of} scarce and noisy resource
\draftadd{availability}, making it a particularly challenging language
for \draftreplace{classical}{most} NLP systems relying on pre-trained
multilingual models \cite{muller2020can}.\footnote{Following
  \citet{seddah-etal-2020-building}, we refer to the Arabizi version
  of North-African Arabic dialects as
  \emph{NArabizi}.} 
To tackle the resource scarcity issue regarding Arabic dialects,
\newcite{antoun2020arabert} use BERT architecture
\cite{devlin-etal-2019-bert} to train a model on Arabic text to
compare this approach to standard multilingual models.  Indeed,
\citet{martin-etal-2020-camemBERT} show that fine-tuning a monolingual
model leads to better results than fine-tuning a multilingual one,
meaning that when fine-tuning is used there is no significant
performance improvement from cross-lingual transfer during
pre-training. However, such model is still pre-trained on sentences
written in a single language and was not trained to handle the
presence of multiple languages in the same sentence (code-switching),
a frequent phenomenon in \textit{NArabizi}.
 
 \draftremove{To tackle this problem, \citet{baert2020arabizi} proposed to pre-train a BERT model on an Egyptian Arabizi data set collected from Twitter. - c pas du tout pour ça qu'ils le font -ds}
 \fi 
 However both monolingual and multilingual model approaches bear the
 risk of being limited by a subword tokenization-based vocabulary
 when facing out-of-domain training data language, especially in
 high-variability  noisy scenarios
 \cite{el-boukkouri-etal-2020-CharacterBERT,clark2021canine}\draftnote{cite(canine-papers)},
 even though \newcite{muller2020can} demonstrated a positive effect
 for \textit{NArabizi} \draftreplace{when adding a certain amount of
   target language data to a multilingual language model in a
   pre-fine-tuning, masked-language model training phase}{when using
   target language data to fine-tune a multilingual language model on
   its own objective function before pre-training}.

 Following a different approach, we investigate the use of a recently
 issued character-based language model
 \cite{el-boukkouri-etal-2020-CharacterBERT} that was shown to display
 a remarkable robustness to lexical variation and noise when facing a
 new distant domain, namely biomedical.  The pipeline we developed is
 simple and consists in fine-tuning this character-based model for
 several tasks in a noisy low-resource language scenario. We show that
 a character-based model trained on only 99k sentences of
 \textit{NArabizi} and fined-tuned on a small treebank of the language
 leads to performance close to that obtained with the same
 architecture pre-trained on large multilingual and monolingual models
 (mBERT and CamemBERT).

 Interestingly, we generalize this observation by using the same
 architecture on a much larger French user-generated Content treebank
 that exhibits similar language variability issues than
 \textit{NArabizi}.  In fact, pre-training a character-based model on
 1\% of the large-scale French instance of the multilingual corpus
 OSCAR leads to similar performance as a subword based model trained
 on the full corpus, showing that such character-based language model
 can reach similar performance levels and that the resulting models
 exhibit the same tolerance to noise as their much larger BERT
 counterparts. This demonstrates the value of such models in very
 scarce resource scenario. Our code and models are freely available.\footnote{\url{https://gitlab.inria.fr/ariabi/character-bert-ugc}}

\section{NArabizi: A Challenging Use Case for NLP in Low-resource Scenarios}\label{sec:NArabizi}
As the official language of 25 countries, Arabic showcases a
linguistic phenomenon called \textit{diglossia} \cite{habash10intro}.
It means that the speakers use Modern Standard Arabic (MSA) for formal
and official situations but use other forms of Arabic in informal
situations. These dialectal forms constitute a dialect continuum with
large variability from one country to the other. Arabic dialects are
defined by their spoken form and often exhibit a lack of standardized
spelling \draftadd{when written}. When Arabic speakers produce written
text in such dialects, they merely transcribe their spoken, colloquial
language, which leads to different forms for the same word. Many users
use the Latin script to express themselves online in their dialect
\cite{seddah-etal-2020-building}. In particular, they transcribe
phonemes that cannot be straightforwardly mapped to a Latin letter
using digits and symbols,\footnote{For example, the digit “3” is often
  used to denote the ayin consonant, because it is graphically similar
  to its rendition in Arabic script.} with a high degree of
variability at all levels; this is called \emph{Arabizi}, with its
North-African version called \textit{NArabizi} by
\newcite{seddah-etal-2020-building}. For cultural and historical
reasons, \textit{NArabizi} also exhibits a high degree of
code-switching with French \draftadd{and Amazigh}
\cite{amazouz2017addressing}, i.e.~alternations between two or more
languages during the conversation. Besides, the only available textual
resources for Arabizi data are user-generated content, which is by
itself inherently noisy
\cite{foster-2010-cba,seddah-etal-2012-french,eisenstein2013bad},
\draftadd{making the production of supervised models, assuming the
  availability of labeled data, or even the collection of large
  pre-training data set a rather difficult
  task.}\draftnote{kaljahi2015foreebank c'est pas foster2010, si ?
  -ds}

This data scarcity problem is often solved in the NLP literature using transfer learning: transferring knowledge learnt by large scale language models pre-trained on larger corpora. However, the use of Latin script makes it harder to transfer knowledge from language models trained on Arabic script corpora.
For subword-based tokenization models, words are represented by a combination of subwords from a predefined list. When applied to a highly variable language with code-switching, a large vocabulary would be necessary to get a good covering of all possible orthographic variations which makes this approach less practical. Table \ref{tab:ex1} presents several examples of lexical variation within \textit{NArabizi}. Interestingly, this variability also affects the
code-switched vocabulary, which is mostly French in this case.

\begin{table}
{\footnotesize
\begin{tabular}{lll}
\toprule
{\sc Gloss} & {\sc Attested forms}               & {\sc Lang}    \\
\midrule
why   & wa3lach w3alh 3alach & {\em NArabizi}  \\ 
all   & ekl kal kolach koulli kol   & {\em NArabizi}  \\
many  & beaucoup boucoup bcp        & Code-switched Fr.  \\ 
\bottomrule
\end{tabular}
\caption{Examples of lexical variation in \textit{NArabizi}. {\em (From \citealp{seddah-etal-2020-building})}}
\label{tab:ex1}
}
\end{table}
\section{Related work}

\subsection{Transfer Learning for Low-resource Languages}

Low-resource languages, by definition, face a lack of textual resources -- annotated or not --, which makes it difficult for the NLP community to develop models and systems adapted to them \cite{joshi-etal-2020-state}. \draftreplace{The majority of exchanges worldwide, whether spoken or written, are actually conducted in one of these languages}{The majority of the almost-7000 languages worldwide actually fall into the ``low-resource'' category}. This makes the development of systems for low-resource languages necessary to widen the accessibility of NLP technology.

For deep learning approaches, which depend on the availability of large data sets, the solution to the low-resource problem comes from the idea of transfer learning. Early instances of cross-lingual transfer learning rely on non-contextualised word embeddings \cite{Ammar2016MassivelyMW}. More recently, multilingual pre-trained language models \cite{conneau-etal-2020-unsupervised} has spread far and wide in NLP, enabling high-performance zero-shot cross-lingual transfer for numerous tasks and languages. The main idea is to exploit a large amount of unlabeled data to pre-train a model using a self-supervised task, such as masked language modeling \cite{lample2019cross,vania2019systematic}. This pre-trained model is then fine-tuned on a much smaller annotated data set and used for another language, domain or task. Strategic knowledge sharing has been shown to improve the performance on downstream tasks and languages \cite{gururangan-etal-2020-dont}. Therefore, this technique is crucial for multilingual applications, as most of the world's languages lack large amount of labeled data  \cite{conneau2019unsupervised, eisenschlos2019multifit,joshi-etal-2020-state}.
However the performance of multilingual language model on low-resource languages is still limited compared to other languages since they are naturally under-sampled during the training process \cite{wu-dredze-2020-languages}.

To improve performance on a specific low-resource languages, there are two possibilities. Either to attempt to train a language model on it from scratch despite the scarceness of data; or fine-tune a pre-trained multilingual language model on the low-resource language corpus, also called \emph{cross-lingual transfer learning} \cite{muller-etal-2021-unseen}. The first option can sometimes lead to decent performance, provided that the training corpus is diverse enough \cite{martin-etal-2020-camemBERT}. 

When following the fine-tuning approach, unsupervised methods can be implemented to facilitate the transfer of knowledge \cite{pfeiffer-etal-2020-mad}. The most widely used unsupervised fine-tuning task is masked language modeling (MLM). This system has proven its efficiency between languages that have already been seen in the training corpus \cite{pires-etal-2019-multilingual}; it is still a challenge for unseen languages, especially low-resource ones. However, \citet{muller2020can} achieved promising results by performing unsupervised fine-tuning on small amounts of \textit{NArabizi} data. We follow their approach by comparing the performance of our pipeline in two setups: {\sc Model+MLM+Task} and {\sc Model+Task}. We describe these setups in more details in section \ref{sec:model+task}. 

\subsection{Tokenization \& Character-based models}

 Standard language models rely on a subword-based approach to process tokens in a sequence \cite{kudo-2018-subword}. This allows the model to handle any word unseen in the training data, working in an ``open-vocabulary setting,'' where words are represented by a combination of subwords from a pre-defined list. On top of \draftreplace{avoiding}{alleviating} the issue of out-of-vocabulary words, this approach allows the model to handle sequences written in a language unseen during training, as long as it uses the same script. Therefore, subword tokenization is a crucial feature of state-of-the-art models in NLP. But its suitability for all types of data has always been questioned. While splitting texts into subwords based on their frequencies works well for English, models using this kind of tokenization struggle with noise, whether it is naturally present in the data \cite{sun2020advbert} or artificially generated to challenge the model \cite{pruthi-etal-2019-combating}. Moreover, language models that use subword-based tokenization struggle to represent rare words \cite{schick2020rare}. Many research projects have focused on improving subword tokenization. For example, \citet{wang2021multiview} suggested a multi-view subword regularization based on the sampling of multiple segmentations of the input text, based on the work of \citet{kudo2018subword}.
 

Other parallel efforts bid on character-based models. For example, \citet{el-boukkouri-etal-2020-CharacterBERT} proposed a possible solution to get a better tokenization system more resilient to orthographic variations and noise in the data set by using a character-level model, inspired by a previous word-level open-vocabulary system \cite{peters-etal-2018-deep}. This new model gets better results than vanilla BERT on multiple tasks from the medical domain. Furthermore, the authors claim that it is more robust to noise and misspellings. In the same vein, \citet{Ma_2020} combined character-aware and subword-based information to improve robustness to spelling errors. This initiated a new wave of tokenizer-free models based on characters or bytes \cite{tay2021charformer,xue2021byt5,clark2021canine}.

The question of knowing if character-based language models can handle high language variability since they are supposed to be resilient to noise and spelling variations is crucial when dealing with non-normalized dialects and non-canonical forms of language as found on many user-generated content platforms. This is why we focus on this work on the analysis of the performance of character-based models on several user-generated content data sets that we now describe.

\section{Data sets}
In this section, we describe the data sets we use to evaluate our pipeline on our downstream tasks, namely Part-Of-Speech (POS) tagging and  dependency parsing.

\paragraph{\textit{NArabizi} Data Set} \label{sec:NArabizitreebank} We
use the \textit{NArabizi} treebank \cite{seddah-etal-2020-building},
containing about 1500 sentences randomly sampled from the romanized
Algerian dialectal Arabic corpus of \citet{cotterell2014algerian} and

from a small corpus of lyrics coming from Algerian dialectal Arabic
songs popular among the younger generation.  This treebank is manually
annotated with morpho-syntactic information (parts-of-speech and
morphological features), together with glosses and code-switching
labels at the word level, as well as sentence-level translations.
Moreover, this treebank also contains 36\% of French tokens. Within
the \textit{NArabizi} annotated
corpus,\footnote{\url{http://almanach-treebanks.fr/NArabizi}} In
addition to labeled data, The \textit{NArabizi} treebanks provides
about \draftadd{2 millions} words\draftadd{, 99k sentences,} of
unlabeled data collected from various sources.\footnote{The original
  data set provides 50k sentences of clean \textit{NArabizi} sentences
  and an additional 49k sentences of more noisy data, we use here a
  concatenation of both.} We use this corpus for unsupervised
fine-tuning.

\paragraph{French Data Sets}
We use the following Universal Dependencies, UD,
\cite{nivre-etal-2020-universal} version of the following treebanks:
French GSD \cite{mcdonald2013universal}, Sequoia
\cite{candito2012corpus} and Spoken, an automatic conversion of the
Rhapsodie corpus \cite{lacheret2014rhapsodie} \draftadd{to the UD
  annotation scheme}.
  
For our experiments on noisy UGC treebank, we use an extension of the
French Social Media Bank, \crapbank\footnote{We use a shuffled version
  of the treebank split into a train set of about 2\,000 sentences and
  a dev and test set of about 1\,000 sentences each.}
\cite{seddah-etal-2012-french}: a treebank of French sentences coming from
various social media sources only available either in constituent
trees or in the native French Treebank dependency annotation scheme
\cite{candito-etal-2010-statistical}, along with the Sequoia original
treebank that we use with the same annotation scheme for compatibility
in our French UGC experiments.  A brief overview of the size and
content of each treebank can be found in Table
\ref{tab:stat_treebanks}.
   
\paragraph{Pre-training Data Sets}
Note that in some of our experiments, we use a fraction, 1\% of the
deduplicated French instance of the Oscar corpora
\cite{suarez2019asynchronous}, about 380M words, as a source of
unlabeled data to be either mixed with \textit{NArabizi} pre-training
data (as an 2.5M words additional sample) or used as \draftreplace{pretraining}{pre-training} data
for CharacterBERT when evaluated on French UGC (whole 1\%). Statistics
on those data sets are presented on Table~\ref{tab:pre_train_stats}.

   \begin{table}
   \centering
\footnotesize
   \scalebox{0.9}{%
\begin{tabular}{lccc}
\hline
Treebank & \# Tokens & \# Sentences & Genres                                                                    \\ \hline
GSD      & 389,363  & 16,342      & \begin{tabular}[c]{@{}l@{}}Blogs, News\\  Reviews, Wiki\end{tabular}       \\ \cline{1-1}
Sequoia  & 68,615   & 3,099       & \begin{tabular}[c]{@{}l@{}}Medical, News\\ Non-fiction, Wiki\end{tabular} \\ \cline{1-1}
Spoken   & 34,972   & 2,786       & Spoken                                                                    \\ \cline{1-1}
\crapbank  &  56,009        &     4,055        & \begin{tabular}[c]{@{}l@{}}Twitter, Facebook\\ Web Forums\end{tabular}    \\ \hline
\end{tabular}%
}
\caption{Statistics on the treebanks used in our POS tagging and
  dependency parsing experiments.}
\label{tab:stat_treebanks}
\end{table}

\begin{table}[h!]
\begin{center}
{\footnotesize
\begin{tabular}{rccc}
\hline
{ } & { \# Tokens} & { \# Sentences} & { Language} \\
\hline
99k Narabizi & 2.527k & 99k & ar-dz \\
\hline
0.01\% Oscar & 3388k & 99k & fr \\
0.1\% Oscar & 33.881k & 993k & fr \\
1\% Oscar & 318.715k & 9.342k & fr \\
10\% Oscar & 1.885.351k & 55.261k & fr \\
100\% Oscar & 23.209.872k & 558.092k & fr \\
\hline
\end{tabular}
\caption{\draftreplace{pretraining}{pre-training} data set statistics.}
\label{tab:pre_train_stats}
}
\end{center}
\end{table}

\draftremove{For our source of pre-training data we use the deduplicated French data set from the OSCAR project \cite{suarez2019asynchronous}. This corpus is extracted from a 2019 Common Crawl Web snapshot, containing more than 23B tokens for this French instance.}

\section{Model}

CharacterBERT \cite{el-boukkouri-etal-2020-CharacterBERT} is a
character-based variant of BERT that replaces the WordPiece embedding
matrix with multiple CNN and a highway layer
\cite{srivastava2015highway}. This method for encoding token
representations is inspired from ELMo \cite{peters2018deep}, one of
the first pre-trained language models for transfer learning. It
generates a context-independent representation from character
embeddings and feeds them to transformer encoder layers, similarly to
BERT architecture. Therefore, it produces a single embedding for any
input token and does not need a WordPiece vocabulary. This avoids
having an inconstant number of subword vectors for each word. We
choose this model since \draftreplace{it is more robustto the noisy
  data we have in our experiments.}{its robustness to noise, when tested
  on biomedical domain by
  \newcite{el-boukkouri-etal-2020-CharacterBERT}, can lead to
  interesting result when facing our noisy experiment data.} It is the first character-based and BERT-like model, along with
CharBERT \cite{ma2020charbert}, that uses both character and subword
embeddings and has the advantage of being publicly
available.\footnote{\url{https://github.com/helboukkouri/character-bert-pre-training}}
Note that we retrain CharacterBert from scratch on our data sets and do not make use of
any of its available  pretrained at models any point in our experiments. 

\section{Experiments}

In this section, we present our fully-supervised and semi-supervised
baselines. We also evaluate different fine-tuning strategies combined
with two layers configurations. We use various embedding models that
we contrast with the CharacterBERT model.

\paragraph{Baseline Models}
For our fully-supervised baseline, we use FastText embeddings
\cite{joulin2016fasttext} trained
from scratch on our \draftadd{treebank} training sets and used
as input for our downstream tasks without any special treatment.  

To measure the effectiveness of using a contextualized character-based
language model, we compare its performance \draftreplace{with}{to}
subword based language models, both monolingual and
multilingual\draftadd{, that constitute the basis of our
  semi-supervised baselines}.  For multilingual subword based
language model, we use mBERT, the multilingual version of BERT
\cite{devlin2018bert}. It is trained for 104 different languages on
Wikipedia data, including French and Arabic, languages for which
\citet{muller2020can} showed that, to a certain extent, they could
transfer to NArabizi. For our monolingual model we use CamemBERT
\cite{martin-etal-2020-camemBERT} which is a contextualized language
model based on the RoBERTa model \cite{liu2019roberta} trained and
optimized specifically for French.

\paragraph{\textbf{\textsc{Model+Task}}} \label{sec:model+task} We use
the implementation of the Biaffine graph parser \cite{dozat2016deep}
from \newcite{grobol:hal-03223424}; they adapted it to handle several
input sources, such as BERT representations. The parser performs its
own tagging using a multi-layer perceptron. The word representations
are a concatenation of word embeddings and tag embeddings learned
together with the model parameters on the treebank training data. \draftadd{We
fine-tune the overall model by back-propagating through the average
of all sub-tokens of a word}. Baseline results, without any external
embeddings, for this parser are provided in Table~\ref{tab:baseline_res}.

\begin{table}
\centering
\footnotesize
\begin{tabular}{lccc}
   & UPOS           & UAS            & LAS             \\ 
\hline
No external embeddings   & 57.61 & 55.48 & 39.32          \\
\hline

\end{tabular}
\caption{Pos-Tagging and Parsing Baseline Results for \textit{NArabizi} test set.}
\label{tab:baseline_res}
\end{table}

\paragraph{\textbf{\textsc{ Model+MLM+Task}}} \label{sec:model+mlm}
Before fine-tuning the model on the downstream task, we perform
language adaptation by fine-tuning it in a self-supervised fashion
with the MLM loss. We use the \textit{NArabizi} raw data; we train the
model for 20 epochs, keeping the best model obtained
\draftreplace{by}{at} the end. The evaluation is done using the MLM
log likelihood, with 10\% of the data kept for validation.

\draftremove{Note that the models that we test without MLM are related to our
test languages, we are testing their cross-lingual ability (French and
Arabic for mBERT models and French for camemBERT)}.

\paragraph{Fine-Tuning Strategy}

In addition to using only the {\em last layer} (cf. Appendix
\ref{appendix:layerconfig} where we conducted several experiments to
explore the effect of different layer configurations in our downstream
tasks), we also test two options for the layers aggregation. The first
one is simply taking the {\em mean} of selected layers.  The second is
{\em scalar mix}, introduced by ELMo~\cite{peters2018deep}: a convex
combination of the transformer layers where the weights are learnt.

\input{table_perf_finetuning_strategy}

For each of these configurations, we test two setup:  {\em with} and {\em without} training the
transformer model weights during the fine-tuning. We call the first
strategy \emph{frozen representation}, noted with a {\em -fz} suffix
in our results, where the language model is used
to extract meaningful features that consist of contextual embeddings.
In the second strategy, the contextual embedding extractor -- that is,
the pre-trained language model -- is fine-tuned on the downstream task
alongside the task-specific component, noted with a {\em -ft} suffix
in our results.

\section{Results and Discussion}
\draftadd{In this section, we compare all the systems presented before on the \textit{NArabizi} and French treebanks. We especially focus on the impact of corpus size on the models' performance.}

 \subsection{Experiments on \textit{NArabizi} treebank}

{\em\footnotesize Scores are reported as triplets describing UPOS/UAS/LAS
    results. Highlighted cells marks the best results column-wise while bold marks best results row-wise}

\iftrue

We report in Table \ref{tab:res_configs} the scores for the different
fine-tuning strategies for the three models CamemBERT, mBERT and
CharacterBERT.

Additional insights on the extraction of a representation from the
different layers are provided in the Appendix, Table \ref{tab:res_layers},
with analysis on the effect of the combination of different subsets of
layers on the accuracy.

\paragraph{CharacterBERT\textsubscript{NArabizi}
  performs better overall in the Model+TASK setting}
Looking at table \ref{tab:res_configs}, we notice that
CharacterBERT\textsubscript{NArabizi} significantly outperforms mBERT
and CamemBERT in the \textsc{Model+Task} setup without MLM for almost
all the configurations, except for the two configurations mean-ft
(when using the mean of all the fine-tuned layers) and scalar-mix-ft
(when using the scalar-mix of all the layers) where mBERT and
CamemBERT show slightly better performance than
CharacterBERT\textsubscript{NArabizi}. In this latter case, mBERT
outperforms CharacterBERT\textsubscript{NArabizi} on the
\textit{NArabizi} data set with accuracy differences of
(3.40/2.13/2.96) for (UPOS/UAS/LAS) using mean-ft, which is only
significant for UAS with p-value<0.05.\footnote{We tested the
  statistical significance using the publicly available Dan Bikel's
  code at 
  \url{https://github.com/tdozat/Parser-v1}} The same observation can
be done for CamemBERT in the {\sc Model+Task} setting which
outperforms CharacterBERT\textsubscript{NArabizi} with accuracy
differences of (-0.14/2.36/2.26). 

 The same is observed for the
scalar-mix-ft option but the differences are not significant with
p-value<0.05 for UAS and LAS. Besides these two configurations,
CharacterBERT\textsubscript{NArabizi} always outperforms the other two
models without MLM. The most notable difference using
CharacterBERT\textsubscript{NArabizi} is when the latter outperforms
CamemBERT with accuracy differences of (19.66/4.39/9.74) and mBERT
with (10.63/3.92/7.854) in the last-layer-fz setting with a
p-value<0.05 for both comparisons.

\paragraph{Adding MLM reverses the trend}
However, when we compare CharacterBERT\textsubscript{NArabizi} to
CamemBERT+MLM and mBERT\discretionary{}{+}{+}MLM, we see that while
they both generally outperform CharacterBERT\textsubscript{NArabizi},
there are some settings where CharacterBERT\textsubscript{NArabizi}
still gets better scores. mBERT+MLM gets the best scores for all of
the configurations we test among the three models. While both
CamemBERT and mBERT perform comparably, mBERT outperforms
CharacterBERT\textsubscript{NArabizi} when it is used with MLM. Some
of the best performance are achieved in the last-layer-ft setting
with scores of (83.08/73.77/62.00) for CamemBERT and
(84.55/73.82/62.67) for mBERT and lower scores of (81.19/70.56/58.6)
for CharacterBERT\textsubscript{NArabizi} (both models outperform
CharacterBERT\textsubscript{NArabizi} in this case).

In other settings, CharacterBERT\textsubscript{NArabizi} is still
competitive with the two other MLM-pretrained models, as illustrated
by the last-layer-fz setting with scores of (70.23/67.86/53.45) for
CamemBERT, and (75.71/66.87/53.69) for
CharacterBERT\textsubscript{NArabizi} and (79.44/71.55/58.98) for
mBERT. {\bf Still, the general tendency in the Model+MLM+TASK setting is that mBERT outperforms
CharacterBERT\textsubscript{NArabizi} when used with MLM and CamemBERT
exhibits performance similar to CharacterBERT\textsubscript{NArabizi}
in the same setting.} This is in contrast with the earlier comparison
without MLM \draftreplace{pretraining}{pre-training} where both CamemBERT and mBERT reached scores
lower than those of CharacterBERT\textsubscript{NArabizi}.

This observation confirms the findings of
\newcite{muller-etal-2021-unseen} regarding the positive impact of
unsupervised fine-tuning for BERT models even if the language is not
one of the pre-training languages.

\paragraph{CharacterBERT pretrained on 1\% of Oscar performs roughly
  the same than
  CamemBERT+Task}  If we compare the performance of CamemBERT in the {\sc Model+Task}
setting to CharacterBERT trained on sub-sample of Oscar, we see that
both models are \draftreplace{competitive in their obtained
  scores}{comparable}.  Typically, CamemBERT seems to outperform
CharacterBERT in some settings like last-layer-ft where the latter
records (78.83/69.52/56.33) for UPOS, UAS and LAS scores respectively
while CamemBERT records higher at (81.14/72.59/60.35). In other
settings, CharacterBERT seems to outperform CamemBERT. In the mean-fz
setting for instance, CharacterBERT has scores of (72.35/66.54/51.98)
which surpasses the (69.47/64.89/49.34) scores of CamemBERT. No clear
conclusion can be drawn about the best use of one model over the other
in the different settings since they all display competitive scores.
This is essentially due to the fact that CamemBERT is trained on the
full Oscar data set, while CharacterBERT is trained on just 0.01\% of
it.  In addition, the test set is made of only 140 \textit{NArabizi}
sentences, making any interpretation of the results difficult. These
two reasons make difficult the  drawing of concrete conclusions on the
performance of both models compared to each other. Therefore in the
next section, we will evaluate the models using the best fine-tuning
strategy on French treebanks with a larger evaluation set and a
CharacterBERT trained on 1\% Oscar.  \fi

\subsection{The Impact of Data Size: Experiments on French treebanks} 

\paragraph{In-domain experiments}
We adopt the best setup from our previous findings for the French treebanks experiments, focusing on the last layer of each model fine-tuned for the task. Table \ref{tab:res_french_treebanks}, which reports scores averaged over five random seeds, compares the performance of Camembert and CharacterBERT on the GSD, SEQUOIA, and SPOKEN treebanks.
Despite being trained on only 1\% of the OSCAR dataset, CharacterBERT performs comparably to Camembert, which is trained on the full dataset. On the GSD treebank, while Camembert slightly outperforms CharacterBERT, the difference is minimal, suggesting that CharacterBERT is highly effective even with significantly less pre-training data. The results for SEQUOIA and SPOKEN exhibit a similar pattern, where Camembert leads by a small margin.

The competitiveness of CharacterBERT across these datasets, especially considering its limited training data, highlights how CharacterBERT can perform strongly despite resource constraints. These results underscore the potential of CharacterBERT in low-resource scenarios, offering comparable performance to models trained on much larger datasets.
\begin{table*}[htb!]
\centering
\scalebox{0.8}{
    \begin{tabular}{lccccccccc}
        \toprule
        \multicolumn{1}{l}{\multirow{2}{*}{Model}} & \multicolumn{3}{c}{GSD} & \multicolumn{3}{c}{SEQUOIA} & \multicolumn{3}{c}{SPOKEN}                                                 \\ \cmidrule{2-10}
                                                   & UPOS                    & UAS                         & LAS                        & UPOS  & UAS   & LAS   & UPOS  & UAS   & LAS   \\ \midrule
        Camembert (100\% Oscar)                                & \textbf{99.32}          & \textbf{95.88}              & 94.70                      & \textbf{99.32} & \textbf{95.88} & \textbf{94.70} & \textbf{97.47} & \textbf{88.21} & \textbf{82.92} \\
        CharacterBERT (1\% Oscar)                              & 99.25                   & 95.12                       & 93.75                      & 99.25          & 95.12          & 93.75         & 96.72          & 86.45          & 80.85          \\
        \bottomrule
    \end{tabular}
}
\caption{POS, UAS and LAS scores on 3 French treebanks, averaged over 5 seeds.}
\label{tab:res_french_treebanks}
\end{table*}

\paragraph{Extremely noisy user-generated content experiments}
In table \ref{tab:res_crapbank}, we compare the performance of Camembert and CharacterBERT on the \crapbank in two different training settings: first, where both models are fine-tuned on the \crapbank training set, and second, where they are trained on the Sequoia treebank.  

In the \crapbank training setting, Camembert and CharacterBERT achieve similar scores, with Camembert having a slight edge on the development set. This trend continues when the models are trained on Sequoia, where both models again show competitive performance. Notably, even though Camembert is pre-trained on the full OSCAR dataset and CharacterBERT only on 1\% of it, CharacterBERT remains highly competitive across both settings, underscoring the effectiveness of its character-based approach in noisy scenarios.

Interestingly, when comparing test set results, the score rankings between the two models are reversed, likely due to random sampling during the dataset split. However, the differences between Camembert and CharacterBERT remain minimal, and the results are not statistically significant, highlighting the comparable performance of both models. Overall, these experiments demonstrate that even with significantly less pre-training data, CharacterBERT performs on par with Camembert when dealing with extremely noisy data. This reinforces the viability of character-based models, especially for user-generated content where variability and noise levels are high, as previously shown in other studies \citep{rosales-nunez-etal-2019-comparison}.
\begin{table*}[htb!]
\footnotesize

	\centering
	\scalebox{0.9}{%
		\begin{tabular}{llcccccc}
			\toprule
		\multicolumn{2}{c}{\multirow{2}{*}{\begin{tabular}[c]{@{}c@{}}Model\end{tabular}}} & \multicolumn{3}{c}{Dev}                  & \multicolumn{3}{c}{Test}                                                                                       \\ \cmidrule{3-8}
		\multicolumn{2}{c}{}                                                               & \rule{0pt}{2.5ex}UPOS                    & UAS                      & LAS            & UPOS           & UAS            & LAS                              \\ \midrule
		\multirow{3}{*}{\rotatebox[origin=l]{90}{\parbox{0.6cm}{\centering  \scriptsize{\crapbank                                                          \\ train}}}}
		& \rule{0pt}{2.5ex}Camembert (100\% Oscar) & \textbf{95.12}           & \textbf{87.03} & \textbf{81.47} & \textbf{95.25} & \textbf{86.91} & \textbf{ 81.87} \\
		& CharacterBERT (1\% Oscar)                & 95.04                    & 86.07          & 80.55          & 95.12          & 86.29          & 81.07           \\ \midrule \midrule
		\multirow{3}{*}{\rotatebox[origin=l]{90}{\textit{\parbox{0.6cm}{\centering \scriptsize{Sequoia                                                                                                                                                   \\train}}}}}   &  \rule{0pt}{2.5ex}Camembert (100\% Oscar)                                      & 89.65    & \textbf{81.41}   & \textbf{74.08}   & 90.35     & \textbf{81.91 }   & \textbf{75.1} \\
		& CharacterBERT (1\% Oscar)                & \textbf{90.08}           & \textbf{81.35} & \textbf{74.04} & \textbf{90.69} & \textbf{81.82} & 74.96           \\  \bottomrule
		\end{tabular}%
	}
	\caption{POS, UAS and LAS scores on \crapbank, averaged over 5 seeds.}
	\label{tab:res_crapbank}
\end{table*}

\section{Discussion}

Our results show that training such a a character-based model
from scratch on much less data gives similar performance to a
multilingual BERT adapted to the language using the same amount of
data for fine-tuning. Overall, our observations confirm the findings of \citet{el-boukkouri-etal-2020-CharacterBERT} regarding the robustness to noise and misspellings of the CharacterBERT model.  We showed that the model has competitive performance on noisy French UGC data when trained on only a fraction of the OSCAR corpus compared to Camembert trained on the full corpus and when trained on corpora containing about 1M words in the extremely noisy and low-resource case of NArabizi. This result is consistent with the findings of \citet{ martin-etal-2020-camemBERT} and \citet{micheli-etal-2020-importance}, who showed that MLM could learn much from pre-training on smaller datasets.  Extending this investigation by training on more data could help explore the model's ability to handle highly variable noisy data.
\begin{table}[htb!]
\footnotesize

	\begin{center}
		{
			\begin{tabular}{rcccc}
				\toprule
				                             & {UPOS}                                                  & {UAS}           & {LAS}           & {\%Oscar} \\
				\midrule
				                             & \multicolumn{4}{l}{\em FSMB fine-tuned (in-domain)}                                                     \\
				{\em No pre-training }       & {\em 81.62}                                             & {\em 69.19}     & {\em 59.17 }    & {\em 0}   \\
				{\em Camembert}              & \textbf{95.25}                                          & \textbf{86.91}  & \textbf{ 81.87} & {\em 100} \\
				Camembert\textsubscript{4gb} & 95.2                                                    & 86.39           & 81.21           & 2.38      \\
				Character-BERT               & 95.12                                                   & 86.29           & 81.07           & 1         \\
				Character-BERT               & 93.78                                                   & 83.13           & 77.49           & 0.1       \\
				Character-BERT               & 91.85                                                   & 79.64           & 73.01           & 0.01      \\
				                             & \multicolumn{4}{l}{\em Sequoia fine-tuned (out domain)}                                                 \\
				{\em No pre-training }       & {\em 72.79}                                             & {\em 59.92}     & {\em 48.81 }    & {\em 0}   \\
				{\em Camembert}              & 90.35                                                   & 81.91           & 75.1            & {\em 100} \\
				Camembert\textsubscript{4gb} & { 90.52}                                                & \textbf{82.25 } & \textbf{75.47}  & 2.38      \\
				Character-BERT               & {\bf 90.69}                                             & 81.82           & 74.96           & 1         \\
				Character-BERT               & 88.33                                                   & 77.28           & 69.89           & 0.1       \\
				Character-BERT               & 85.81                                                   & 73.29           & 65.07           & 0.01      \\
				\bottomrule
			\end{tabular}
			\caption{CharacterBert model performance compared with a small
				Camembert (4gb) model on the \crapbank test set in in-domain and
				out-of-domain fine-tuning scenarios. {\em Full-size Camembert
						results are reported here for reference.}}
			\label{tab:4gbvsChar}
		}
	\end{center}
\end{table}

However, one could question the usefulness of such CharacterBERT models if small BERT-based models were available in the same domain. To build an answer to that question, we conducted a quick set of experiments comparing our CharacterBERT model trained on 1\% of OSCAR with the off-the-shelf Camembert version trained on 4 GB of the OSCAR corpus French instance (2.38\% of the entire corpus), Camembert\textsubscript{4GB}, and which was shown to perform almost as well as the full model \citep{ martin-etal-2020-camemBERT} on many downstream tasks. Both models were fine-tuned according to our \modeltask architecture on either the \crapbank or the Sequoia treebank, allowing us to evaluate their in-domain and out-of-domain performance. Results on Table~\ref{tab:4gbvsChar} confirm the effectiveness of our CharacterBERT model with very close results to Camembert\textsubscript{4gb} in the in-domain scenario and similar, if not slightly better in the out-of-domain scenario in POS tagging. The very good performance of Camembert\textsubscript{4GB} raises some interesting questions. The fact that Camembert\textsubscript{4GB} was trained on more than twice as much data and with 100k pre-training steps while the CharacterBERT pre-training stopped below 20k steps probably explains this small discrepancy, but further investigations are needed with a fully parallel setting where both CharacterBERT and Camembert are trained on the same amount of data and the same hyper-parameters. Therefore, we trained CamemBERT and CharacterBERT models from scratch on a new 4GB sample from the French OSCAR \textit{21.09} corpus using identical settings and matching training steps to build a more accurate comparison. The resulting new version of CamemBERT\textsubscript{4gb} is called Camembert\textsubscript{4gb}V2. This parallel setup allows us to directly evaluate the performance of each model under equivalent conditions fine-tuned and tested on \crapbank. The results in Table \ref{tab:4gbnew} reveal that CharacterBERT, even with identical training conditions and data size, slightly outperforms Camembert\textsubscript{4gb}V2 across UPOS, UAS, and LAS metrics. This observation supports our hypothesis that CharacterBERT’s character-based architecture provides a competitive advantage in managing linguistic variability, making it especially suitable for low-resource and highly variable language scenarios. 
\begin{table}[htb!]
\footnotesize

	\begin{center}
		{
			\begin{tabular}{rcccc}
				\toprule
				                             & {UPOS}                                                  & {UAS}           & {LAS}           & {\%Oscar} \\
				\midrule

				 CamemBERT\textsubscript{4gb}V2          & 94.65                                          & 84.99  & 79.6 & 2.38 \\
				CharacterBERT\textsubscript{4gb} & \textbf{94.98}                                                     & \textbf{86.45}           & \textbf{81.28}           & 2.38      \\                             
				\bottomrule
			\end{tabular}
			\caption{Performance comparison of CharacterBERT and CamemBERT models trained on the same 4GB sample of French OSCAR , fine-tuned and tested on \crapbank .}
			\label{tab:4gbnew}
		}
	\end{center}
\end{table}   

The take-home message from these experiments is that CharacterBERT seems to efficiently capture at least some of the UGC idiosyncracies compared to its BERT-based counterparts, given its small pre-training data size. This finding was also shown by \citet{RosalesNunes_et_al:2021:NoisyUGCTranslation} in the context of character-based neural machine translation. Interestingly, their results showed that transformer-based models with subword tokenization also exhibit strong robustness to a particular type of lexical noise. This behavior has also been demonstrated by \citet{ItzhakLevy:2021:SpellingBee} and could explain why the BERT-based models we tested performed so well in our experiments. The key seems to be relying on the ability of the subword distribution to model some forms of lexical variations. More experiments are needed to investigate in what circumstances, in addition to noisy and resource-scarce scenarios, CharacterBERT models bring a notable advantage. 

Our results are based on the evaluation of two low-level tasks. Therefore, it would be interesting to see if they can be generalized to other tasks, e.g., more semantic, as additional experiments on model layers configuration showed that most of the important information is captured early in the model's layers.

Regarding Arabic dialects written in Arabizi, a recent BERT-based model has been trained on 7 million Egyptian tweets and showed effective results in a sentiment analysis task \citep{baert-etal-2020-arabizi}. Contemporary to the development of CharacterBERT for \narabizi, another model was pre-trained on 4 million Algerian tweets and demonstrated interesting results on sentiment analysis \citep{Abdaoui_et_al:2021:DziriBERT}. Unfortunately, the authors did not perform any experiments on the Narabizi data set, thus making the comparison with our work not straightforward. However, we believe that given the shortcomings of finding enough data to train large models for dialects, and based on our findings concerning the training set size,  it would probably also be promising to improve the quality of the training and fine-tuning datasets for a more accurate evaluation.

In summary, We showed that CharacterBert models trained on very little data could
provide an interesting alternative to large multilingual and
monolingual models in resource-scarce and noisy scenarios. This is why
we release all the code, data and models to reproduce our experiments,
hoping  our work will favor the rise of efficient robust NLP models
for under-resourced languages, domains and dialects.\footnote{\url{https://gitlab.inria.fr/ariabi/character-bert-ugc}}

\section*{Acknowledgments}
We thank the reviewers for their very valuable feedback.  The first author was partly funded by 
Benoît Sagot's chair in the PRAIRIE institute funded by the French
national agency ANR as part of the ``Investissements d’avenir''
programme under the reference ANR-19-P3IA-0001. This work also  
received funding from the European Union’s Horizon 2020 research and
innovation programme under grant agreement No. 101021607 and from the
French Research Agency via the ANR ParSiTi project (ANR-16-CE33-0021).

\bibliography{anthology,custom}
\bibliographystyle{acl_natbib}

\input{Appendix}

\end{document}

%% file: draft_final.tex
\expandafter\def\expandafter\UrlBreaks\expandafter{\UrlBreaks
  \do\a\do\b\do\c\do\d\do\e\do\f\do\g\do\h\do\i\do\j%
  \do\k\do\l\do\m\do\n\do\o\do\p\do\q\do\r\do\s\do\t%
  \do\u\do\v\do\w\do\x\do\y\do\z\do\A\do\B\do\C\do\D%
  \do\E\do\F\do\G\do\H\do\I\do\J\do\K\do\L\do\M\do\N%
  \do\O\do\P\do\Q\do\R\do\S\do\T\do\U\do\V\do\W\do\X%
  \do\Y\do\Z}

\usepackage{soulutf8}
\usepackage[normalem]{ulem}
\usepackage{xcolor}

\usepackage{enumitem}
\setlist{nolistsep}
\usepackage{microtype}

\iftrue
    \usepackage[final]{proofing}
  \renewcommand\hl[1]{{#1}}  
   \newcommand\draftHL[2]{#1}{\draftnote{\red{#2}}}
   \newcommand\redHL[1]{}
  \newcommand\todo[1]{}

  \newcommand{\Djame}[1]{}

\else  

\usepackage[draft]{proofing}

\newcommand{\Djame}[1]{
\textbf{\textcolor{red}{\hl{Djame: #1}}}
}

\newcommand\red[1]{{{\textcolor{red}{\bf #1}}}}

\newcommand\draftHL[2]{\hl{#1}{\draftnote{\red{#2}}}}

\let\oldred\red
\renewcommand\red[1]{{ \oldred{{#1}}}}

 \renewcommand\draftHL[2]{\hl{#1}{\draftnote{\red{#2}}}}
 \newcommand\redHL[1]{\red{\hl{#1}}}
\let\olddraftnote\draftnote
\renewcommand\draftnote[1]{\olddraftnote{\red{#1}}}

\fi

%% file: table_perf_finetuning_strategy.tex
\begin{table*}
\centering
\begin{subtable}[t]{0.495\linewidth}
\centering
\footnotesize
\scalebox{0.9}{%
\begin{tabular}{@{}lcccccc@{}}
\multirow{2}{*}{\parbox{1.5cm}{\centering Fine-tuning Strategy}} & \multicolumn{3}{c}{\textsc{Model+Task}}                                               & \multicolumn{3}{c@{}}{\textsc{Model+MLM+Task}}                                            \\ 
\cline{2-7}
{\rule{0pt}{2.5ex}} & {UPOS} & {UAS} & {LAS} & {UPOS} & {UAS} & {LAS}  \\ 
\hline
 last-layer-ft & \textbf{81.45} & \textbf{71.7}  & 59.83 & \cellcolor{bananamania}{\textbf{83.85}} & \cellcolor{bananamania}{\textbf{74.01}} & \cellcolor{bananamania}{\textbf{62.93}}\\
                        last-layer-fz & 59.76 & 62.1  & 44.2  & \cellcolor{bananamania}{70.99} & \cellcolor{bananamania}{67.32} & \cellcolor{bananamania}{52.72} \\
                        mean-ft & 79.01 & 72.09 & 59.39 & \cellcolor{bananamania}{80.8}  & \cellcolor{bananamania}{73.03} & \cellcolor{bananamania}{60.9}  \\
                        mean-fz & 65.18 & 63.17 & 46.64 & \cellcolor{bananamania}{73.47} & \cellcolor{bananamania}{68.26} & \cellcolor{bananamania}{54.56} \\
                        scalar-mix-ft & 80.28 & \textbf{71.76} & \textbf{58.93} & \cellcolor{bananamania}{82.86} & \cellcolor{bananamania}{73.55} & \cellcolor{bananamania}{61.77} \\
                        scalar-mix-fz  & 65.89 & 63.63 & 47.54 & \cellcolor{bananamania}{73.24} & \cellcolor{bananamania}{68.26} & \cellcolor{bananamania}{54.35} \\
\end{tabular}%
}
\caption{ \textbf{CamemBERT}}
\label{tab:res_Camembert_oscar_configs}
\end{subtable}
\hfill
\begin{subtable}[t]{0.495\linewidth}
\centering
\footnotesize
\scalebox{0.9}{%
\begin{tabular}{@{}lcccccc@{}}
\multirow{2}{*}{\parbox{1.5cm}{\centering Fine-tuning Strategy}} & \multicolumn{3}{c}{\textsc{Model+Task}}                                               & \multicolumn{3}{c@{}}{\textsc{Model+MLM+Task}}                                            \\ 
\cline{2-7}
{\rule{0pt}{2.5ex}} & {UPOS} & {UAS} & {LAS} & {UPOS} & {UAS} & {LAS}  \\ 
\hline
                        last-layer-ft  & \cellcolor{bananamania}{69.39}     & \cellcolor{bananamania}{60.32}    & 48.04    & 68.01      & 59.84     & \cellcolor{bananamania}{50.22} \\
                        last-layer-fz & 64.27     & 63.21    & 46.72    & \cellcolor{bananamania}{77.65}      & \cellcolor{bananamania}{71.13}     & \cellcolor{bananamania}{58.79}     \\
                        mean-ft & 68.58     & 64.84    & 49.42    & \cellcolor{bananamania}{72.34}      & \cellcolor{bananamania}{68.55}     & \cellcolor{bananamania}{53.98}     \\
                        mean-fz  & 69.56     & 64.71    & 50.35    & \cellcolor{bananamania}{79.56}      & \cellcolor{bananamania}{71.65}     & \cellcolor{bananamania}{59.47}     \\
                        scalar-mix-ft & \textbf{81.14}     & \textbf{70.03}    & \textbf{58.37}    & \cellcolor{bananamania}{\textbf{84.75}}      & \cellcolor{bananamania}{\textbf{73.27}}     & \cellcolor{bananamania}{\textbf{62.32}}     \\
                        scalar-mix-fz & 69.62     & 65.24    & 50.86    & \cellcolor{bananamania}{79.4}       & \cellcolor{bananamania}{71.62}     & \cellcolor{bananamania}{59.21}     \\
\end{tabular}%
}
\caption{\textbf{mBERT}}
\label{tab:res_mebert_oscar_configs}
\end{subtable}\\
\vspace{0.2cm}
\begin{subtable}{\linewidth}
\footnotesize
\centering
\scalebox{0.9}{%

\begin{tabular}{lccccccccc|ccc}
\multirow{3}{*}{\parbox{2cm}{\centering Fine-tuning Strategy}} & \multicolumn{3}{c}{NArabizi}                                                 & \multicolumn{3}{c}{Sample Oscar}                                             & \multicolumn{3}{c}{NArabizi + Oscar}  & \multicolumn{3}{|c}{Sample Oscar + MLM NArabizi}                                        \\ 
\cline{2-13}
{\rule{0pt}{2.5ex}} & \multicolumn{3}{c}{99k}                                                      & \multicolumn{3}{c}{99k}                                                      & \multicolumn{3}{c}{66k+33k} & \multicolumn{3}{|c}{99k+99k}                                                  \\ 
\cline{2-13}
{\rule{0pt}{2.5ex}} & {UPOS} & {UAS} & {LAS} & {UPOS} & {UAS} & {LAS} & {UPOS} & {UAS} & {LAS} & {UPOS} & {UAS} & {LAS}  \\ 
\hline
                        last-layer-ft & \textbf{80.42} & 69.86 & 57.69 & 79.21 & 68.39 & 55.99 & \textbf{80.72} & \textbf{69.39} & \textbf{57.13}  & \cellcolor{bananamania}{\textbf{82.48}} & \cellcolor{bananamania}{70.25} & \cellcolor{bananamania}{58.37}\\
                        last-layer-fz & 72.16 & 66.41 & 52.36 & 68.6 & 64.86 & 50.2 & 73.29 & 67.93 & 53.57 & \cellcolor{bananamania}{77.02} & \cellcolor{bananamania}{68.57} & \cellcolor{bananamania}{55.58} \\
                        mean-ft & \textbf{80.44} & \textbf{70.34} & \textbf{58.22} & \textbf{79.82} & 68.72 & 56.12 & 79.29 & 69.13 & 56.32 & \cellcolor{bananamania}{82.36} & \cellcolor{bananamania}{70.55} & \cellcolor{bananamania}{\textbf{59.02}}\\
                        mean-fz & 75.45 & 67.43 & 54.61 & 71.11 & 65.99 & 51.42 & 76.76 & 68.94 & 55.92 & \cellcolor{bananamania}{79.17} & \cellcolor{bananamania}{70.22} & \cellcolor{bananamania}{57.74} \\
                        scalar-mix-ft  & 79.49 & 69.84 & 57.97 & 79.53 & \textbf{68.95} & \textbf{56.24} & 77.92 & 69.27 & 56.25 & \cellcolor{bananamania}{82.39} & \cellcolor{bananamania}{\textbf{70.66}} & \cellcolor{bananamania}{58.5} \\
                        scalar-mix-fz & 72.16 & 66.41 & 52.36 & 68.6 & 64.86 & 50.2 & 73.29 & 67.93 & 53.57 & \cellcolor{bananamania}{77.02} & \cellcolor{bananamania}{68.57} & \cellcolor{bananamania}{55.58} \\               
\end{tabular}%
}

\caption{\textbf{CharacterBERT} (\textsc{Model+Task})}
\label{tab:res_charcterbert_configs}
\end{subtable}
\caption{Performances of the models on the \textit{NArabizi} treebank using \textbf{different fine-tuning strategy} (We use \emph{ft} to indicate that the embeddings are fine-tuned for the tasks, while \emph{fz} is used when the embeddings are frozen during the fine-tuning step).}
\label{tab:res_configs}
\end{table*}

%% file: Appendix.tex
\appendix

 \section{Layer Configuration Experiments}\label{appendix:layerconfig}

We focus on the scalar mix strategy to study the effect of the combination of different subsets of layers on the accuracy of the downstream task, instead of only aggregating all layers. For example, it has been shown that higher layers contain more semantic information, while lower layers contain more syntactic information \cite{jawahar2019does}. We compare several layer configurations, that is different subsets of the transformer from which we get the sentence representation.

\input{table_per_layer_configuration} 
 
 \paragraph{The effect of the Layer Configuration}
We report in tables \ref{tab:res_Camembert_Oscar_layers}, \ref{tab:res_mBert_layers} and \ref{tab:res_charcterbert_layers} the scores for the different layers combinations for CamemBERT, mBERT and CharacterBERT respectively.  
For mBERT and CamemBERT, the performance increases when using the last layers, while for CharacterBERT, there is no big difference between the different layers combinations. For example for the setup {\sc Model+Task}, the accuracy for POS tagging (UPOS) goes from around 70 with the first layer using CamemBERT and mBERT, and reaches 80 using CharacterBERT\_arabizi. For CharacterBERT, these scores stay around 80 even when using farther layers (for layers 4 to 7 for instance, the UPOS score is 80.53 and for layers 6-11 it is around 80.34), and while using the last layer gives the best score of 81.19, the latter is still considered around 80 and the stagnation in the scores is hence visible. Contrarily, for CamemBERT and mBERT, the UPOS scores for the {\sc Model+Task} setup increase from around 70 using only the first layer to above 80 when using layers 6 through 11 (80.39 for CamemBERT and 80.72 for mBERT). The best UPOS score for CamemBERT appears when using the last layer alone (81.14) while for mBERT it is when using layers 6 through 11 (80.72). This clearly illustrates the increase \draftreplace{in the scores}{of performance} when using higher layers for CamemBERT and mBERT. The same observation can be made for UPOS scores in the {\sc Model+MLM+Task} setup for mBERT and CamemBERT, and for the other Unlabeled Attachment Score (UAS) and Labeled Attachment Score (LAS) scores as well.

One possible explanation is that the information captured by CharacterBERT layers \draftreplace{is the same}{does not evolve} along the model's layers. The model produces a single embedding for any input token based on an aggregation of the characters embeddings while for BERT-like-models, each sub-word unit in a word is embedded using a WordPiece embedding matrix. Therefore, a possible interpretation is that CharacterBERT learns all the information at the earliest layers as we feed it the whole word directly and not an inconstant count of sub-words when WordPiece vocabulary is in use. Moreover, less than 10\% of the 100\,000 most frequent \draftreplace{words}{sub-words} in the \textit{NArabizi} raw data are present in the mBERT vocab due to the high variability of NArabizi.

%% file: table_per_layer_configuration.tex
\begin{table*}
\centering
\footnotesize
\begin{subtable}{0.48\linewidth}
\centering
\scalebox{0.9}{%
\begin{tabular}{lcccccc}
\multirow{2}{*}{\parbox{1cm}{\centering Layer Config}} & \multicolumn{3}{c}{\textsc{Model+Task}}                                               & \multicolumn{3}{c@{}}{\textsc{Model+MLM+Task}}                                            \\ 
\cline{2-7}
{\rule{0pt}{2.5ex}} & {UPOS} & {UAS} & {LAS} & {UPOS} & {UAS} & {LAS}  \\ 
\hline
 0 & {68.38} & \cellcolor{bananamania}{66.21} & \cellcolor{bananamania}{50.61} & \cellcolor{bananamania}{70.89} & {64.74} & {49.81} \\
                        0-5 & {77.93} & {70.51} & {57.04} & \cellcolor{bananamania}{80.95} & \cellcolor{bananamania}{72.21} & \cellcolor{bananamania}{59.45} \\
                        4-7 & {79.63} & {70.18} & {57.56} & \cellcolor{bananamania}{81.57} & \cellcolor{bananamania}{72.54} & \cellcolor{bananamania}{60.92} \\
                        6-11 & {80.39} & {71.88} & {58.41} & \cellcolor{bananamania}\textbf{83.65} & \cellcolor{bananamania}{74.15} & \cellcolor{bananamania}{62.15} \\
                        11 & \textbf{81.14} & \textbf{72.59} & \textbf{60.35} & \cellcolor{bananamania}{83.08} & \cellcolor{bananamania}{73.77} & \cellcolor{bananamania}{62.00} \\
                        all & {80.15} & {70.46} & {58.08} & \cellcolor{bananamania}\textbf{83.65} & \cellcolor{bananamania}\textbf{74.62} & \cellcolor{bananamania}\textbf{62.62} \\     
\end{tabular}
}
\caption{\textbf{CamemBERT}}
\label{tab:res_Camembert_Oscar_layers}
\end{subtable}\hfill
\begin{subtable}{0.48\linewidth}
\centering
\footnotesize
\scalebox{0.9}{%
\begin{tabular}{lcccccc}
\multirow{2}{*}{\parbox{1cm}{\centering Layer Config}} & \multicolumn{3}{c}{\textsc{Model+Task}}                                               & \multicolumn{3}{c@{}}{\textsc{Model+MLM+Task}}                                            \\ 
\cline{2-7}
{\rule{0pt}{2.5ex}} & {UPOS} & {UAS} & {LAS} & {UPOS} & {UAS} & {LAS}  \\ 
\hline
                        0 & {73.63} & {65.36} & {52.08} & \cellcolor{bananamania}{75.43} & \cellcolor{bananamania}{66.21} & \cellcolor{bananamania}{52.98} \\
                        0-5 & {79.40} & \textbf{69.19} & {57.23} & \cellcolor{bananamania}{82.37} & \cellcolor{bananamania}{69.90} & \cellcolor{bananamania}{58.08} \\
                        4-7 & {80.62} & {69.61} & {57.33} & \cellcolor{bananamania}{83.74} & \cellcolor{bananamania}{72.45} & \cellcolor{bananamania}{62.05} \\
                        6-11 & \textbf{80.72} & {68.53} & {56.76} & \cellcolor{bananamania}\textbf{85.02} & \cellcolor{bananamania}{72.40} & \cellcolor{bananamania}{61.63} \\
                        11 & {80.48} & \textbf{69.19} & \textbf{57.89} & \cellcolor{bananamania}{84.55} & \cellcolor{bananamania}\textbf{73.82} & \cellcolor{bananamania}\textbf{62.67} \\
                        all & {80.25} & \textbf{69.19} & {56.14} & \cellcolor{bananamania}{84.31} & \cellcolor{bananamania}{72.78} & \cellcolor{bananamania}{61.67} \\
\end{tabular}
}
\caption{\textbf{mBERT}}
\label{tab:res_mBert_layers}
\end{subtable}\\
\begin{subtable}{\linewidth}
\footnotesize
\centering
\begin{tabular}{lccccccccc}

\multirow{3}{*}{\parbox{1cm}{\centering Layer Config}} & \multicolumn{3}{c}{NArabizi}                                                 & \multicolumn{3}{c}{Sample OSCAR}                                             & \multicolumn{3}{c}{NArabizi + Oscar}                                          \\ 
\cline{2-10}
{\rule{0pt}{2.5ex}} & \multicolumn{3}{c}{99k}                                                      & \multicolumn{3}{c}{99k}                                                      & \multicolumn{3}{c}{66k+33k}                                                   \\
\cline{2-10}
{\rule{0pt}{2.5ex}} & {UPOS} & {UAS} & {LAS} & {UPOS} & {UAS} & {LAS} & {UPOS} & {UAS} & {LAS}  \\
\hline
0 & 78.92  & \cellcolor{bananamania}\textbf{70.79} & \cellcolor{bananamania}57.99 & 78.26 & 69.61 & 56.76 & \cellcolor{bananamania}79.25 & 69.33 & 57.14 \\
0-5 & \cellcolor{bananamania}79.63 & \cellcolor{bananamania}70.23 & \cellcolor{bananamania}57.99 & 79.54 & 70.13 & 56.76  & 79.06 & 68.05 & 55.25 \\
4-7 & \cellcolor{bananamania}80.53  & 69.28 & 57.47 & 79.77 & \textbf{70.75} & \cellcolor{bananamania}\textbf{57.84}  & 80.10  & \cellcolor{bananamania}\textbf{69.90} & 56.76 \\
6-11 & 80.34 & \cellcolor{bananamania}70.60 & \cellcolor{bananamania}57.84 & 78.26 & 68.43 & 55.58 & \cellcolor{bananamania}\textbf{80.91} & 69.66 & \textbf{57.80} \\
11  & \cellcolor{bananamania}\textbf{81.19}  & \cellcolor{bananamania}70.56 & \cellcolor{bananamania}\textbf{58.65} & 78.83 & 69.52 & 56.33 & 80.67 & \textbf{69.90} & 57.75  \\
all & 79.96 & 69.47 & \cellcolor{bananamania}57.84 & \cellcolor{bananamania}\textbf{80.62} & 68.76 & 56.76 & 80.15 & \cellcolor{bananamania}\textbf{69.90} & 57.56                   
\end{tabular}

\caption{ \textbf{CharacterBERT} }
\label{tab:res_charcterbert_layers}
\end{subtable}
\caption{Performances of the models on the \textit{NArabizi} treebank using \textbf{different combinations of the model layers} for embeddings.}
\label{tab:res_layers}
\end{table*}